\documentclass{article}

\usepackage[preprint]{neurips_2020}

\usepackage[utf8]{inputenc} 
\usepackage[T1]{fontenc}    
\usepackage{hyperref}       
\usepackage{url}            
\usepackage{booktabs}       
\usepackage{amsfonts}       
\usepackage{nicefrac}       
\usepackage{microtype}      
\usepackage{graphicx}
\usepackage{wrapfig}
\usepackage{caption}

\title{FireBERT: Hardening BERT Classifiers Against Adversarial Attack}

\author{
  Gunnar Mein, Kevin Hartman, Andrew Morris\\
  UC Berkeley MIDS students\\
  \texttt{gunnarmein@berkeley.edu, kevin.hartman@berkeley.edu, andrewmorris@berkeley.edu}
}

\begin{document}

\maketitle

\begin{abstract}
We present FireBERT, a set of three proof-of-concept NLP classifiers hardened against TextFooler-style word-perturbation by producing diverse alternatives to original samples. In one approach, we co-tune BERT against the training data and synthetic adversarial samples. In a second approach, we generate the synthetic samples at evaluation time through substitution of words and perturbation of embedding vectors. The diversified evaluation results are then combined by voting. A third approach replaces evaluation-time word substitution with perturbation of embedding vectors. We evaluate FireBERT for MNLI and IMDB Movie Review datasets, in the original and on adversarial examples generated by TextFooler. We also test whether TextFooler is less successful in creating new adversarial samples when manipulating FireBERT, compared to working on unhardened classifiers. We show that it is possible to improve the accuracy of BERT-based models in the face of adversarial attacks without significantly reducing the accuracy for regular benchmark samples. We present co-tuning with a synthetic data generator as a highly effective method to protect against 95\% of pre-manufactured adversarial samples while maintaining 98\% of original benchmark performance. We also demonstrate evaluation-time perturbation as a promising direction for further research, restoring accuracy up to 75\% of benchmark performance for pre-made adversarials, and up to 65\% (from a baseline of 75\% orig. / 12\% attack) under active attack by TextFooler.
\end{abstract}

\section{Introduction}

\subsection{Motivation and prior work}

Just as we have seen interesting, easy-to-fabricate but hard-to-explain adversarial attacks against visual classifiers \citep{ma2020understanding, eykholt2018robust}, attacks against text classification systems have been proposed. Earlier examples use misspellings \citep{li2018textbugger}, an attack easily thwarted with preprocessing \citep{pruthi2019combating} or co-training \citep{zhu2019freelb}. 

In 2019, a group of researchers from MIT, University of Hong Kong, and A*STAR Singapore published a new approach, creating perturbations focused on flipped classification but minimal semantic difference. The “TextFooler” project \citep{jin2019bert} used word similarities to generate adversarial corpora prevailingly understood and classified the same way as the original by humans. The technique fools BERT near completely on many benchmarks, exposing a potentially dangerous vulnerability to adversarial attack. TextFooler performs a near exhaustive search for adversarial samples that come reasonably close to preserving the semantic content of the original. It treats the classifier as a black box, examining only the classifier probability output to determine the importance of individual words. 

Recent work to guard against such attacks \citep{karimi2020adversarial} co-trains a BERT model on embedding perturbations generated by following the gradients. It achieves small but measurable success. It picks up earlier work \citep{goodfellow2014explaining} which explored (among other methods) adding a defined noise distribution to input vectors. Goodfellow’s team concluded that adversarial samples are not finely distributed around input vectors, but, rather, that there are “pockets” of  adversarial classification, and that ensembles will offer no protection against carefully constructed adversarial samples. Their work was based on MNIST classification, but their claim is that the findings apply more generally to most networks that feature large linear components. We claim that BERT-based NLP classifiers are sufficiently removed from simple linear behavior to allow us to take another look at the effectiveness of ensembles.

TextFooler specifically attacks BERT, not its descendants. Thus our goal is to evaluate our work against TextFooler directly. We know of no paper attempting to harden against an attack by TextFooler or any adversary with similarity to its exhaustive approach and degree of success. There are numerous BERT derivatives, and some claim added robustness \citep{liu2019roberta}. They may offer additional resilience above regular BERT against adversarial attack, but, to our knowledge, this has not been evaluated. The TextFooler paper itself has been recently updated but the code remains unchanged. Updates to the paper were confined to the analysis and future work suggestions, and are not relevant to our result. A recent survey of adversarial attacks and defenses \citep{wang2019survey} does not include TextFooler and makes the claim that adversarial attacks against text are impractical; a statement which TextFooler’s achievements should put very much in doubt. 

\section{Primary contribution}

MNLI entailment and IMDB sentiment classification accuracy with BERT under adversarial attack by purely pre-made TextFooled samples (generated with TextFooler on previously known BERT-based classifiers) can be improved from approximately 0\% to close to original performance with a hardened classifier. TextFooler analysis of (and sample generation on) such a hardened classifier can be made significantly more difficult (requiring at least 5 times the amount of computation performed for an attack on an unmodified BERT-based classifier, and failing with at least twice the unmodified rate). We show that this can be achieved without substantially lowering the regular accuracy for the above-named benchmarks.

Our principal contribution consists of three classifiers constructed as a defense mechanism against TextFooler-style attacks, with details discussed below. The code, plus tuned models, hyperparameter search code and training and evaluation notebooks for these classifiers, in addition to tools for exploratory data analysis and the actual training and evaluation data, are available at our anonymous GitHub \href{https://github.com/FireBERT-author/FireBERT}{\underline{repository}}. A summary of our most important results:

\begin{itemize}
\item Reducing the error rate on pre-made adversarial samples by 79\% (new accuracy 0.800) on MNLI and 87\% (new accuracy 0.872) on IMDB by co-tuning with synthetic samples.
\item Reducing the error rate by 62\% on both the MNLI and IMDB tasks (to 0.623 and 0.620, respectively) on the IMDB tasks through evaluation-time vector perturbation.
\item Reducing the error rate by 48\% (to accuracy 0.545) on the MNLI task under active TextFooler attack, through evaluation-time vector perturbation.
\item We show that TextFooler overfits to a specific, tuned model: Simply re-tuning on the original data improves accuracy against pre-made adversarial samples significantly.
\end{itemize}

\subsection{Methods}

We explore three ways to teach BERT to be more accepting of perturbed sentences while preserving classification results. All three are applied to both sentiment classification (IMDB) and entailment classification (MNLI): In approach 1 (“FuSE”), we introduce additional, slightly word-diversified samples during the evaluation, and make a voting ensemble. In approach 2 (“FIVE”), we shortcut the search for replacement words and add Gaussian noise to the input-embedding vectors directly. In approach 3 (“FACT”), we co-tune the classifier with the same diversified samples we use in FuSE. All three use a shared component to perturb text, which we will explore in more detail. All three classifiers are built around an underlying BERT-instance. They are implemented as subclasses of the base classifier, modifying only very specific parts of the behavior.

\pagebreak

\subsubsection{SWITCH - "Substituting Words In Text Classification Hardening"}

\begin{wrapfigure}[36]{r}{0.55\textwidth}
   \centering
   \vspace{-10pt}
   \includegraphics[width=0.55\textwidth]{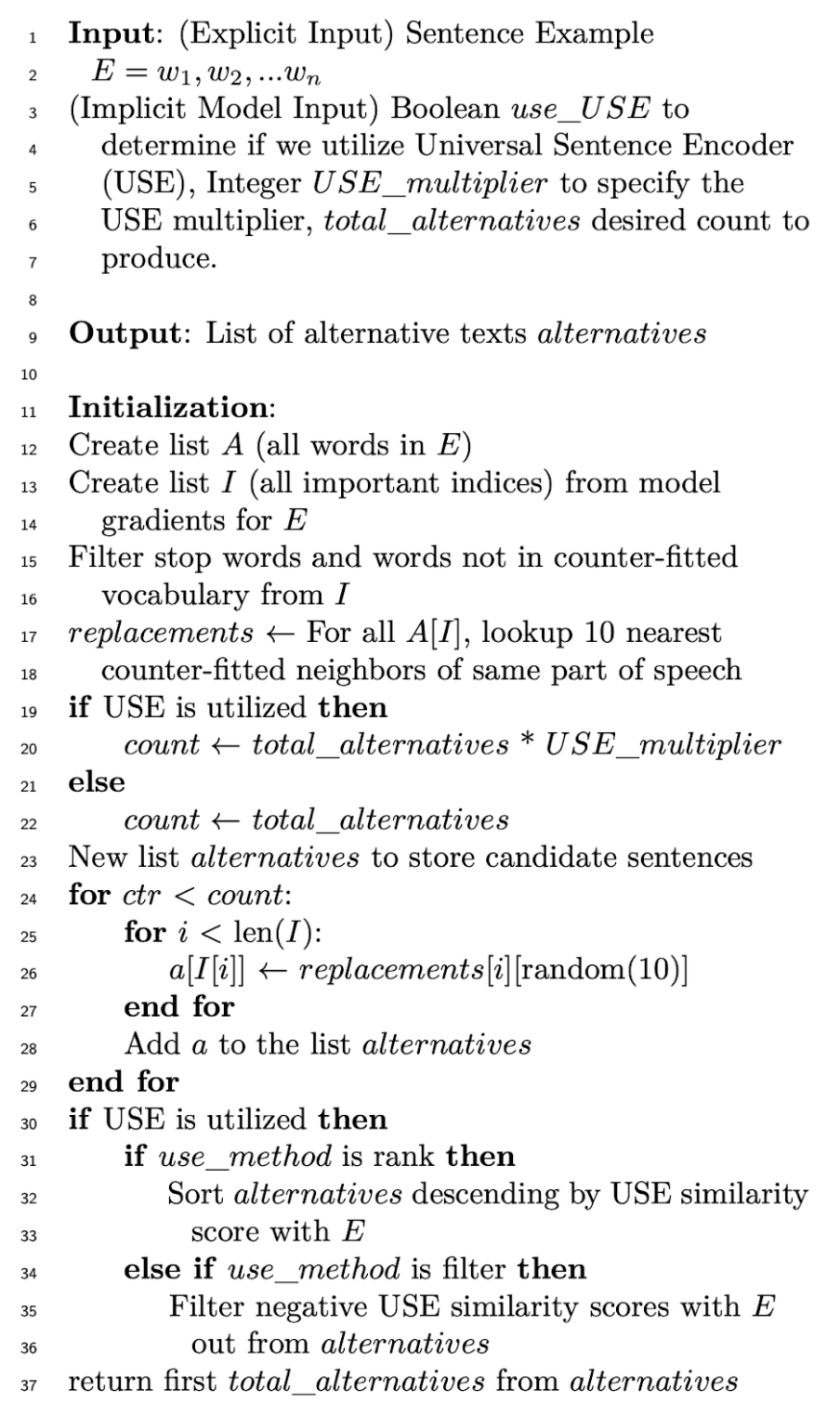}
    \caption{SWITCH}
    \label{fig:pseudocode}
\end{wrapfigure}

The purpose of our SWITCH component is to provide sample diversity with retained classification. SWITCH takes an example like this: “this movie is truly fun for the whole family adults and kids will totally enjoy it!” and produces alternatives like this: “this photography is sincerely fun for the whole family matures and teenagers will perfectly enjoy it !" and “this theatre is truly fun for the whole family forties and kiddies will entirely enjoy it !”.

SWITCH uses its own pre-tuned BERT instance for evaluating the gradients for the input sentence/pair provided, in order to determine which words are important to the classification. It then uses the same counter-fitted embeddings employed by TextFooler to create alternative words. The actual cosine similarities are never needed, since words far away from the original are of no interest to us. We store a pre-computed matrix of 100 nearest-neighbor index numbers for each word. Replacement words are filtered through part-of-speech matching, and finally the replacement texts can optionally be ranked (for closest semantic similarity to the original) or filtered (for at least positive similarity value) through the Universal Sentence Encoder (USE) \citep{cer2018universal} similarity scores. There is a random element to SWITCH’s final choice of alternatives, in order to make it harder for the exhaustive trial-and-error process of TextFooler to have a stable target to work with. Tunable hyperparameters include the number of words to perturb, the number of alternative samples to generate, whether to use part-of-speech matching, and whether to employ USE in either filtering or ranking, plus a multiplier to generate more samples before USE is applied.

\subsubsection{TextFooler baseline models}

To establish a baseline for our three approaches we obtain the original models provided by the TextFooler authors which were fine-tuned on bert-base-uncased for the MNLI and IMDB tasks. These pre-tuned models are the BERT instances fed into SWITCH for querying and active searching of diverse candidates.

\subsubsection{Secondary Pytorch Lightning baseline models}

The base code for our three classifiers is a reimplementation of a HuggingFace BERT-based uncased sequence classifier in Pytorch Lightning. We use the published TextFooler binary models for baseline results. We fine-tune secondary baseline models for the IMDB and MNLI tasks, to validate our code and the training parameters. After a random hyperparameter search, we select 5 training epochs and a batch size of 32 (MNLI) and 20 (IMDB) with a learning rate of $2*10^{-5}$ and no weight decay. Adam epsilon is maintained consistently at $1*10^{-8}$. 

\pagebreak

\subsubsection{Fuzzy sentence ensemble - FuSE}

\begin{wrapfigure}[28]{l}{0.60\textwidth}
    \centering
    \vspace{-10pt}
    \includegraphics[width=0.60\textwidth]{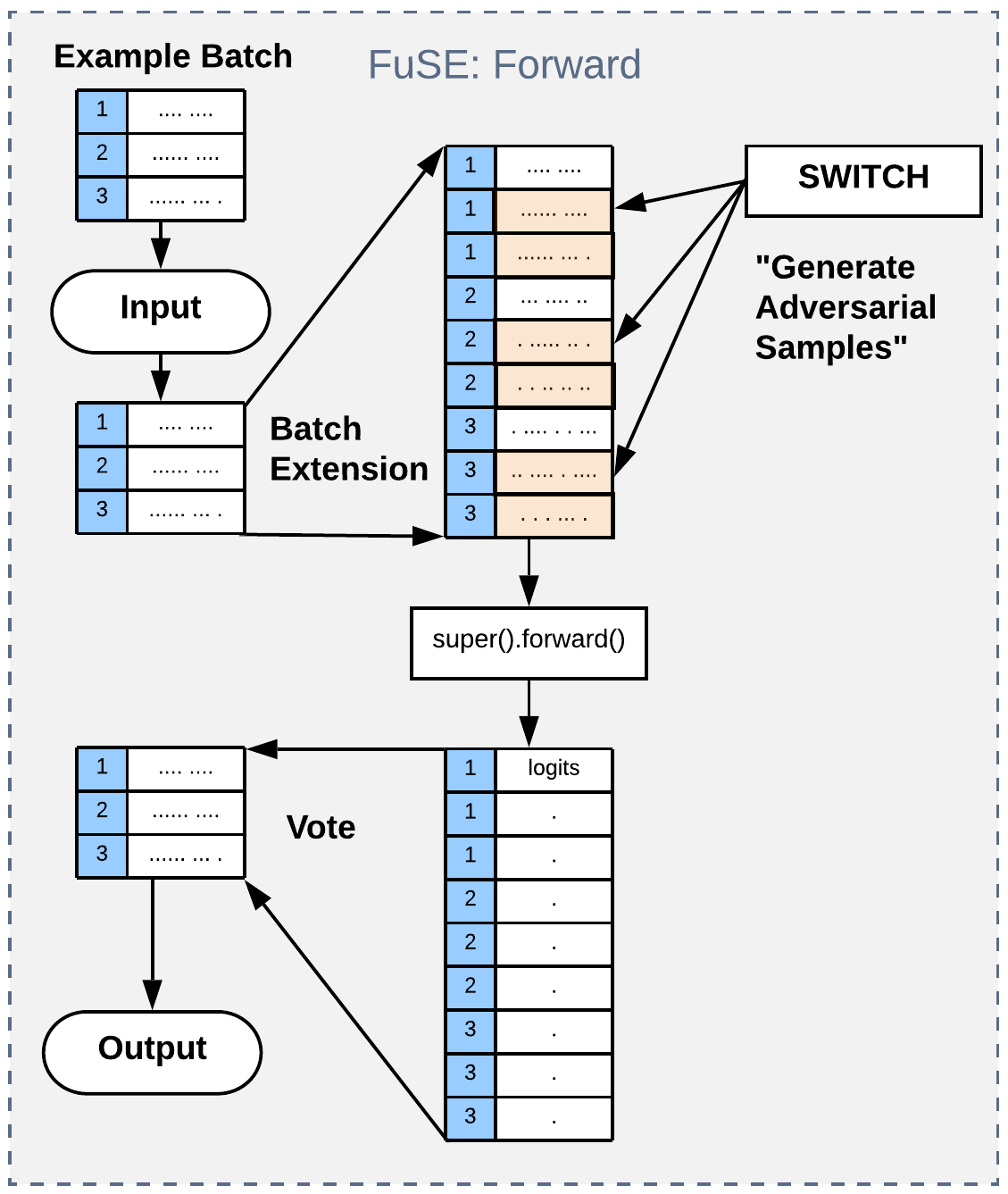}
    \caption{FuSE perturbation method}
    \label{fig:fuse}
\end{wrapfigure}

The hypothesis for FuSE is that sampling nearest neighbors for a word will provide, on average, better classification outputs. In other words, most neighboring words will provide correct classification rather than adversarial classification results. FuSE uses SWITCH to determine the most important words to the classification through gradient computation. After determining the important words, SWITCH provides alternative sample formulations by changing those words with a number of neighboring words. Using this method, FuSE assembles a random number of alternative sample sentences and evaluates all of them against the underlying sentiment or entailment classifier, as seen in figure \ref{fig:fuse}. FuSE outputs either synthetic logits representing a majority vote count of classifications for the samples ("majority vote"), or the average of the logits across the samples ("logit-averaging"). Tunable hyperparameters are all SWITCH parameters, plus the selection of the voting method.
 
\subsubsection{Fuzzy internal vector ensemble - FIVE}

\begin{wrapfigure}[21]{r}{0.65\textwidth}
    \centering
    \vspace{-10pt}
    \includegraphics[width=0.65\textwidth]{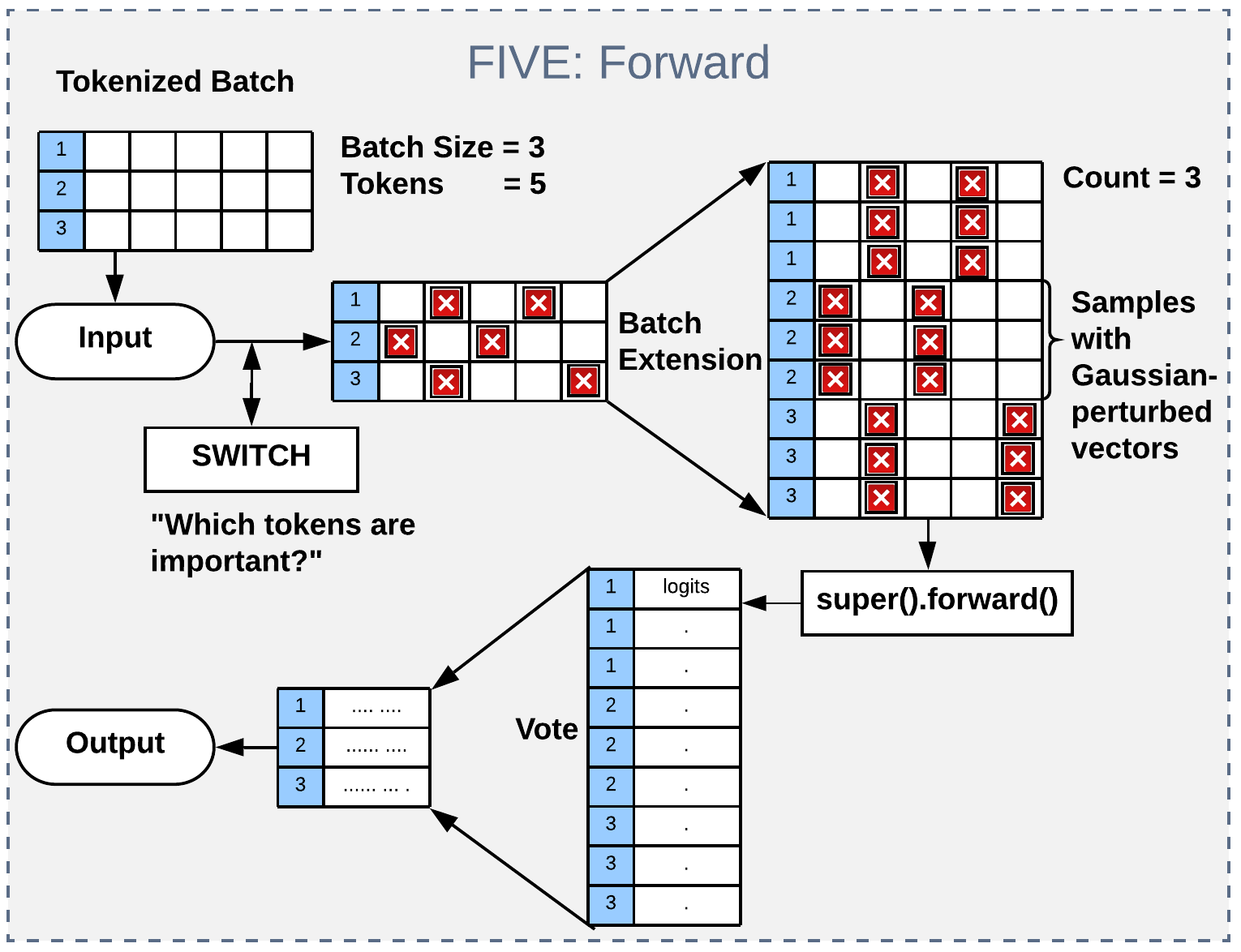}
    \caption{FIVE perturbation method}
    \label{fig:five}
\end{wrapfigure}

FIVE is based on the hypothesis that averaging over neighborhoods of embedding vectors, evaluated in the context of the sample, leads to more stable average classifications than evaluating based on a particular embedding that might have been changed by an adversary like TextFooler. FIVE asks SWITCH to identify the most important word by gradient computation, and then creates additional synthetic samples by perturbing their token embedding vectors, as seen in figure \ref{fig:five}. Each set of perturbed embeddings forms a Gaussian distribution around their original vector. Like FuSE, FIVE outputs either synthetic logits representing a vote count of classifications for the samples ("majority vote"), or the average of the logits across the samples ("logit-averaging"). Tunable hyperparameters include the number of embeddings to perturb, number of perturbed samples to generate, the standard deviation of the Gaussian distribution to create around the original embeddings, and the voting method for combining the individual synthetic sample votes. 

\pagebreak

\subsubsection{Fuzzy adversarial co-tuning - FACT}

\begin{wrapfigure}[16]{l}{0.60\textwidth}
    \centering
    \vspace{-10pt}
    \includegraphics[width=0.60\textwidth]{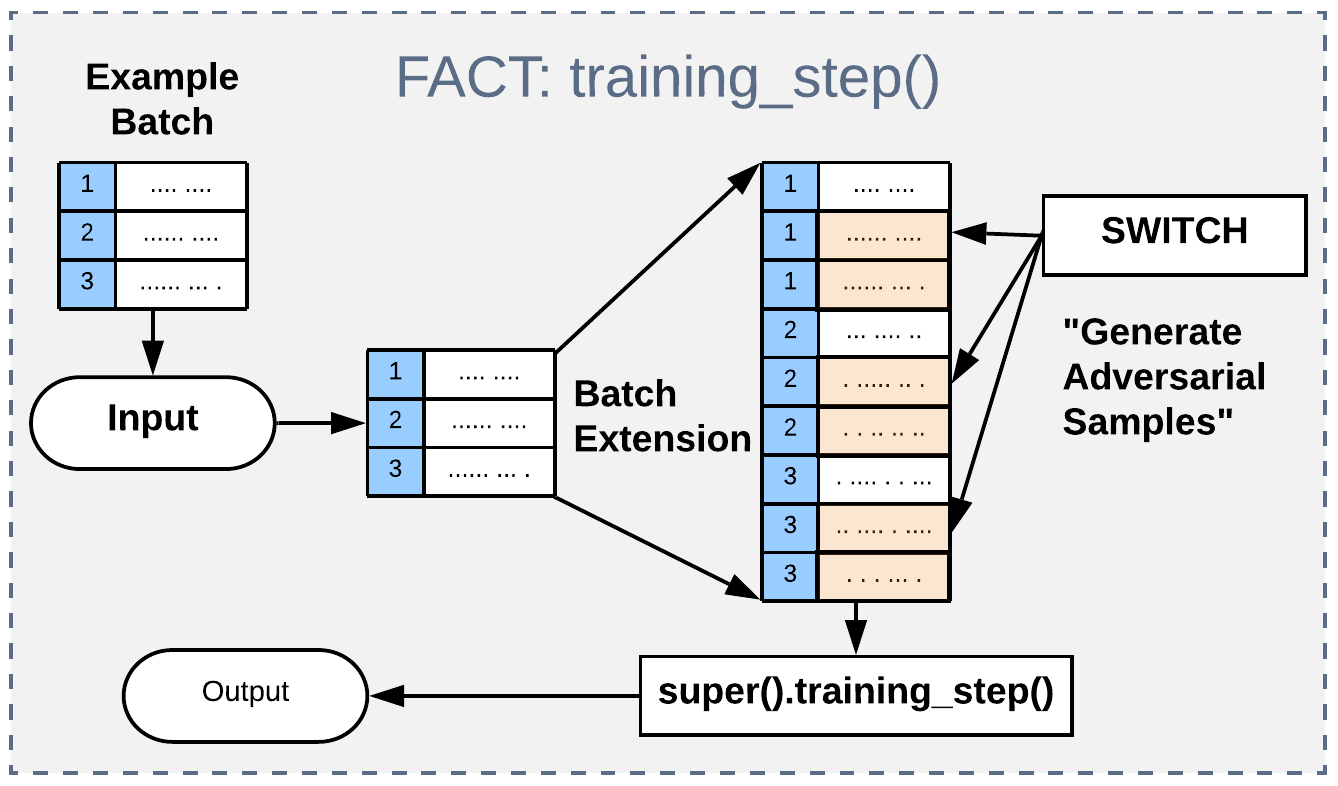}
    \caption{FACT in-batch co-tuning}
    \label{fig:fact}
\end{wrapfigure}

In our third, fine-tuning-based approach, we introduce a little language diversity to the tuning process to teach the FACT classifier that there are multiple ways to express our original sentiment or entailed fact. In this way, we are fine-tuning BERT to become less sensitive to the very specific adversarial examples TextFooler generates. FACT uses SWITCH to identify important words and provide alternative samples, which are injected sequentially inside the batch during the fine-tuning process (Figure \ref{fig:fact}). The advantage of this approach is that, after fine-tuning, we can just deploy a new binary model file for evaluation or use in production. On the downside, co-tuning with synthetic samples can substantially increase the fine-tuning time depending on the number of additional samples requested. But this is performed only once, and the resulting model has no additional run-time degradation. All SWITCH hyperparameters are tunable, and regular training hyperparameters like batch size, learning rate, weight decay and Adam epsilon can also be adjusted.  

All three classifiers inherit from our Pytorch Lightning base classifier which provides hooks for batch extension with perturbed samples during the forward() (for FuSE and FIVE) and training\_step() (for FACT) methods. The classifiers override these extension hooks to provide diversified samples in their respective approaches.

\section{Results and conclusions}

\subsection{Evaluation procedures and source data}

For IMDB training and evaluation, we download 50,000 IMDB labeled movie review samples \citep{kaggle2019movie} and split the data into 40,000, 5,000 and 5,000 samples for train, validation and test. The data was originally curated by Stanford University \citep{maas2011learning}

For MNLI \citep{williams2017broad} training and evaluation, we download data from the GLUE baseline repository \citep{wang2018glue}. Training is performed on the provided file of 390K samples. Validation is performed on the 10,000 samples from dev\_matched. Since labels are not publicly available in the test set, we use the 10,000 samples from dev\_mismatched as our holdout for testing.

We generate adversarial samples for each of the train, validation and test sets in source data by running the TextFooler algorithm on each of the base models for our complete data sets. The adversarial data derived from validation samples are used for hyperparameter tuning and evaluation of our three approaches. The adversarial samples derived from test data are reserved for single use in final evaluation. We provide these adversarial sets for each of the tasks to further efforts in this research.

Performance under active attack by TextFooler is evaluated using the code, metrics and datasets provided by TextFooler. A set of 1000 samples for the MNLI task was selected by the TextFooler authors. TextFooler runs masked versions of the samples through the classifiers to establish which words most affect the outcome. It then runs the classifier on samples with those words perturbed to nearest neighbors until it finds good adversaries, sometimes unsuccessfully.  

\subsection{Hyperparameter search}

Hyperparameter search for FIVE hyperparameters, and the SWITCH parameters of FuSE and FACT, is performed in a time-boxed fashion with a random search of the hyperparameter space. Random searches were utilized based on demonstrated benefits with this method \citep{bergstra2012random}.

\pagebreak

\subsection{Results}

\begin{wraptable}[12]{l}{0.65\textwidth}
  \centering
  \begin{tabular}{lllll}
    \toprule
    Model          & Acc.           & F1 score       & Adv. acc.      & Adv. \\
                   &                &                &                & F1 score\\
    \midrule
    Baseline (TF)  & \textbf{0.843} & \textbf{0.835} & 0.029          & 0.027 \\
    Baseline (New) & 0.833          & 0.826          & 0.501          & 0.487 \\
    FIVE           & 0.757          & 0.708          & 0.632          & 0.566 \\
    FuSE           & 0.725          & 0.713          & 0.591          & 0.568 \\
    FACT           & 0.827			& 0.821          & \textbf{0.800} & \textbf{0.791} \\
    \bottomrule \\
    \multicolumn{5}{c}{\textbf{Table 1: MNLI results}} \\
  \end{tabular}
\end{wraptable}

Table 1 shows the results of all classifiers against the MNLI dataset. The original MNLI model provided by TextFooler serves as the baseline. Its performance against TextFooler samples that were generated specifically against it is unsurprisingly low. The secondary (freshly tuned) baseline model is slightly inferior to the TextFooler baseline mode on originals, but brings up the accuracy on adversarial samples substantially. FIVE achieves a substantial improvement in the adversarial case at the expense of significantly, but not unreasonably, dampened performance on the originals (1 perturbed embedding, std dev 8.14, 8 synthetic samples per original, logit-averaging). For FuSE, we find similar results with slightly worse performance on original samples (2 perturbed words, 10 candidates per word, part-of-speech matching, 14 candidates into USE, filter negative scores, max 14 samples, logit-averaging). Co-tuning with FACT results in the best performance against adversarials while sacrificing barely any accuracy on originals (batch size 7, 9 words to perturb, 10 candidates per word, part-of-speech matching, 12 candidates into USE, filter negative scores, max 4 samples).

\begin{wraptable}[12]{r}{0.61\textwidth}
  \centering
  \begin{tabular}{lllll}
    \toprule
    Model          & Acc.           & F1 score       & Adv. acc.      & Adv. \\
                   &                &                &                & F1 score\\
    \midrule
    Baseline (TF)  & \textbf{0.906} & \textbf{0.904} & 0.002          & 0.002 \\
    Baseline (New) & 0.905          & 0.902          & 0.827          & 0.816 \\
    FIVE           & 0.884          & 0.867          & 0.620          & 0.586 \\
    FuSE           & 0.518          & 0.508          & 0.778          & 0.770 \\
    FACT           & 0.900          & 0.897          & \textbf{0.872} & \textbf{0.867} \\
    \bottomrule \\
    \multicolumn{5}{c}{\textbf{Table 2: IMDB results}} \\
  \end{tabular}
\end{wraptable}

Table 2 shows the results against the IMDB dataset. The original TextFooler IMDB model serves as the baseline. The secondary (freshly tuned) baseline model shows itself to be basically not vulnerable to pre-manufactured  adversarial samples - retuning addresses the problem by itself for this dataset. FIVE achieves significant gains in the adversarial case while not losing substantial accuracy on the benchmarks (1 perturbed embedding, std dev 2.3, 10 synthetic samples per original, probability averaging). For FuSE, we find good performance against adversarials, but a complete degradation to coin-flip level for the original samples.  More work is needed to understand this result  (3 perturbed words, 10 candidates per word, part-of-speech matching, 17 candidates into USE, filter negative scores, max 12 samples, probability averaging). Co-tuning with FACT once again performs well against both original and adversarial samples (batch size 2, 23 words to perturb, 10 candidates per word, part-of-speech matching, 12 candidates into USE, filter negative scores, max 4 samples).

\begin{wraptable}[14]{l}{0.35\textwidth}
  \centering
  \begin{tabular}{lll}
    \toprule
    Model          & Org.          & Adv.          \\
                   & Acc.*         & Acc.*         \\
    \midrule
    Baseline (TF)  & 0.851         & 0.127         \\
    FUSE           & 0.499         & 0.276         \\
    FIVE           & 0.777         & \textbf{0.463}\\
    FACT           & 0.820         & 0.316         \\
    FuSE (FACT)    & 0.373         & 0.373         \\
    FIVE (FACT)    & 0.743         & \textbf{0.545}\\
    \bottomrule \\
    \multicolumn{3}{c}{\textbf{Table 3: Accuracy }} \\
    \multicolumn{3}{c}{\textbf{under active attack}} \\
  \end{tabular}
\end{wraptable}

We also investigate how well our classifiers perform against an active attack by TextFooler. This required a minimal adaptation of TextFooler to work against our Pytorch Lightning classifiers. Fully explaining TextFooler’s result parlance is beyond the scope of this paper, but briefly (Table 3, all numbers generated by TextFooler code, and adversarial accuracy "Adv. Acc*" not directly comparable to tables 1 and 2): In our baseline measurement, TextFooler degrades the accuracy of a BERT sequence classifier to around 12\%. We find that FACT is able to raise that number to a significant 31\% on MNLI, requiring TextFooler to change around 30\% more words and try about 30\% more samples. FuSE on top of a FACT-tuned model raises the number again. Against FIVE, with no re-tuning, TextFooler uses about the same number of perturbed words and classifier queries but is unable to degrade the accuracy below 45\%. FIVE on top of the co-tuned FACT model delivers the best performance of all at 54.5\% accuracy.

\pagebreak

\subsection{Analysis}

\begin{wrapfigure}[22]{r}{0.58\textwidth}
    \centering
    \vspace{-10pt}
    \fbox{\includegraphics[width=0.58\textwidth]{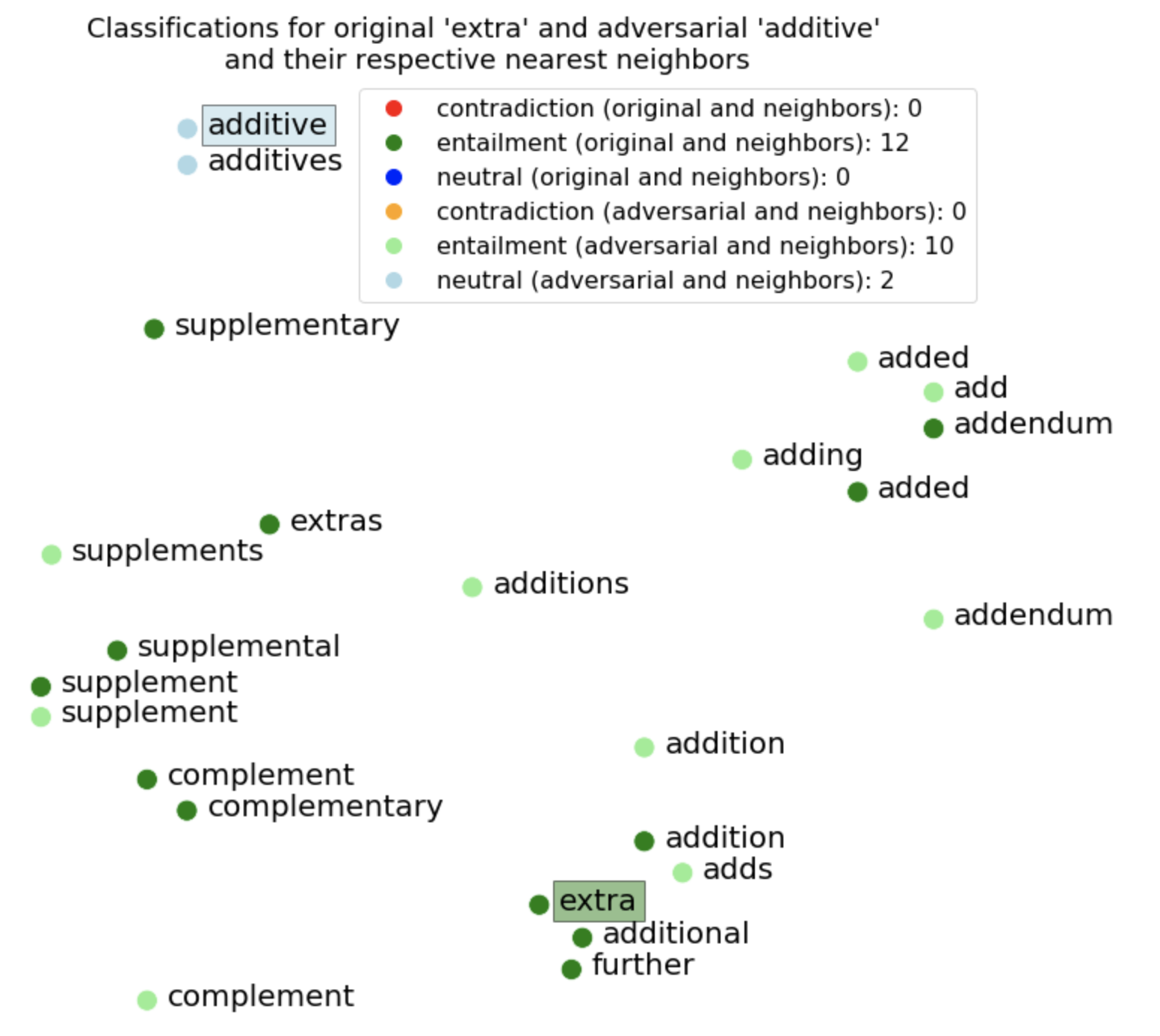}}
    \caption{Nearest neighbors of words (MNLI baseline)}
    \label{fig:plotfuse}
\end{wrapfigure}

A look at a “fooled” MNLI example can be instructive in understanding what is going on inside our classifiers. Here is one from our validation set: \{Premise: “So I have to find a way to supplement that.”, Hypothesis: “I need a way to add something extra.”, Label: "entailment"\}. TextFooler is able to minimally change the hypothesis in a way that most of us would reasonably still classify the same way, fooling the classifier into a “neutral” classification: \{Hypothesis: "I need a way to add something additive."\}. To look at what happens in the evaluation-time classifiers FuSE and FIVE, we will perturb only one of the words for illustration. By computing the gradients for the classification and finding the input that has the largest absolute gradient, SWITCH correctly decides that “extra” and “additive” are the most important words for the respective original and adversarial hypotheses. 
 
\begin{wrapfigure}[23]{l}{0.59\textwidth}
    \centering
    \vspace{-10pt}
    \fbox{\includegraphics[width=0.58\textwidth]{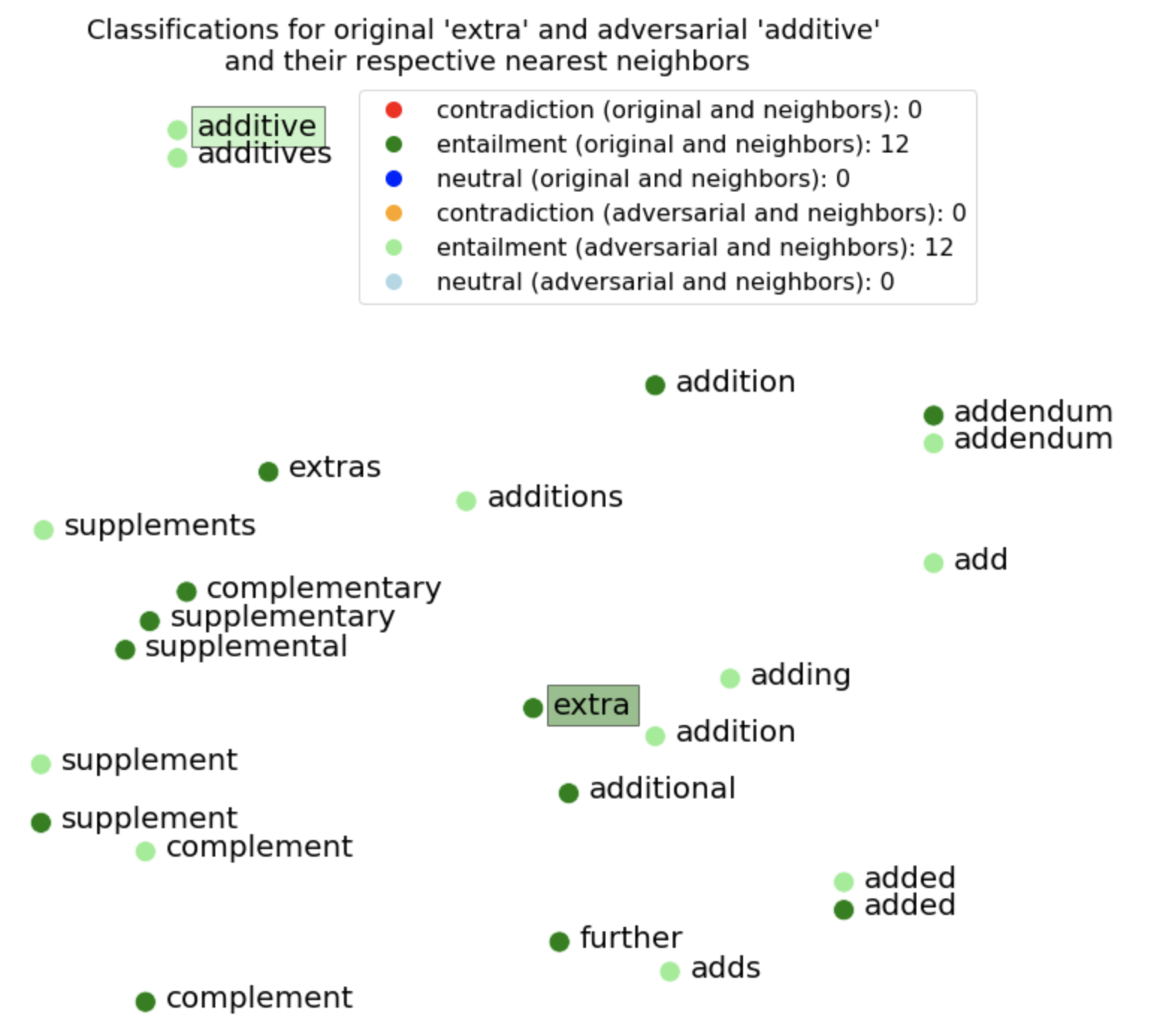}}
    \caption{Nearest neighbors of words (MNLI FACT)}
    \label{fig:plotfact}
\end{wrapfigure}

Figure \ref{fig:plotfuse} shows classification outcomes for the replacement words evaluated in the context of their respective full premise/hypothesis pairs. The nearest neighbors (by cosine similarity) around the original word "extra" all lead to an "entailment" classification. Among its neighbors, the adversarial word "additive" is one of only two leading to a "neutral" classification, with the majority landing on "entailment". This is the kind of example we had hoped to see - a stable neighborhood of words around the original, but a majority of original classifications around the adversarial sample. TextFooler found one isolated word, the smallest possible nudge to give the sample, to change the classification. If we look at the neighborhood as a whole and average across it, we find a more stable classification. FuSE won’t be fooled  by TextFooler in this example.

For co-tuning (FACT), we ask SWITCH to come up with a list of diverse replacement hypotheses: "i need a method to additions something additive", "i need a pathway to additions something other", "i need a manner to inserts something additional", "i need a manner to totals something add". More points of stable classification are established with the BERT-model. Tuned on this extra set of samples, FACT no longer considers the substitution of “additive” adversarial (figure \ref{fig:plotfact}). 

We chose a particularly benign sample to illustrate the workings here - not all examples work out this well. MNLI defense in particular is a difficult problem to solve at evaluation time, as it is often easy to nudge a sample from “entailment” or “contradiction” to “neutral” with a single word perturbation, but overwhelmingly unlikely to reverse that judgment with another random nudge.

\pagebreak

\begin{wrapfigure}[19]{r}{0.58\textwidth}
    \centering
    \fbox{\includegraphics[width=0.58\textwidth]{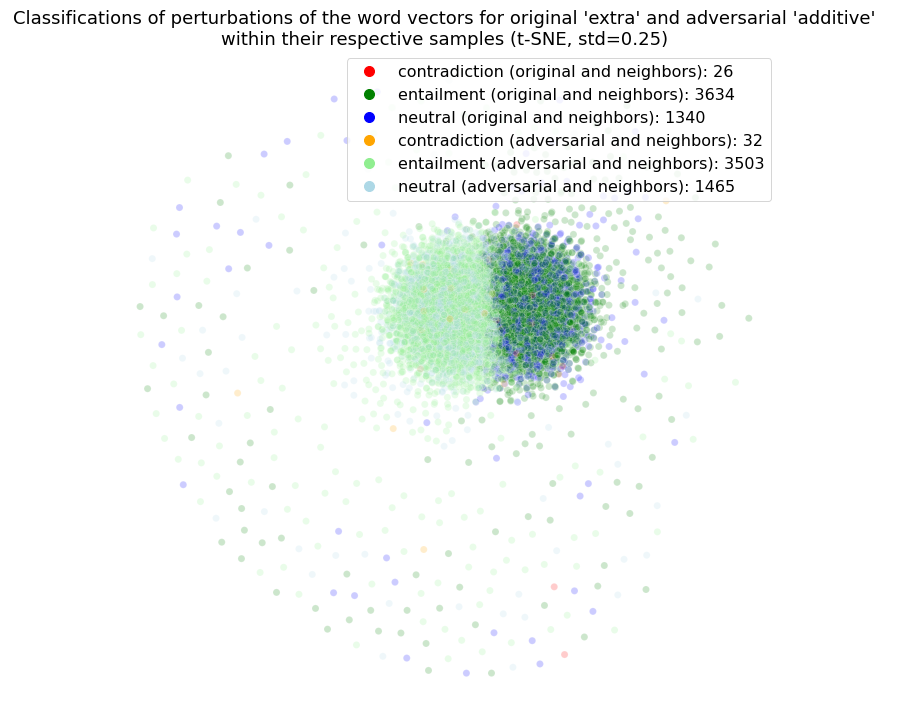}}
    \caption{Classification fields (FIVE vector perturbation)}
    \label{fig:plotvecs}
\end{wrapfigure}

From the substitution of words, we now switch into the embedding space that contains the inputs to the actual BERT-classifier. Figure \ref{fig:plotvecs} shows the classification fields for vector perturbation of the most important words in our samples, at a standard deviation of 0.25. Note that this is not the best hyperparameter for classification with FIVE, but it shows the clustering well in this t-SNE diagram. The Gaussian regions around both the original and the adversarial words are laced with adversarial points, and at a small standard deviation, we will indeed find regions of predominantly adversarial sentiment. With the right tuning, however, we find a standard deviation hyperparameter at which the adversarial classification represents only small pockets, and as we are averaging over the cluster, FIVE comes to the right conclusion for this example as well.

\subsection{Discussion}

Our first insight is that TextFooler overfits to the model it works against when generating samples. We find that adversarial samples generated from one model (e.g. the original MNLI model from  \citet{jin2019bert}) will fool a second model (e.g. the secondary MNLI model tuned on PyTorch Lightning) in only half the circumstances. The same results are also true in the reverse (e.g. samples generated from the Lightning model will fool the original model only half the time). This leads us to suspect that TextFooler works to find the weak points in the specific way a model is parameterized.

FuSE delivers a solid performance on the MNLI task, and fails spectacularly on regular (non-adversarial) IMDB samples, achieving nothing more significant than coin-flip performance in the most “conservative” of our hyperparameter searches. More investigation is needed to understand this failure, especially since its performance against pre-made adversaries is substantial at 77\% accuracy.

We find that co-tuning with SWITCH is a very effective way to protect a model against TextFooler’s original samples. We also see that evaluation-time perturbation can improve adversarial results, trading in a few degrees of task accuracy for a good degree of adversarial protection. Additionally, active attack query-count results show that the co-tuned models get harder to fool in generating new sets of samples for MNLI, but not for IMDB. We suspect this is because the sentence length is much larger in IMDB than in MNLI, and therefore gives TextFooler more possible word choices in perturbation.

However, our most resilient model against active attack is the FIVE classifier, which preserves accuracy to 45\% (55\% on top of FACT). We theorize that it is the stronger random aspect of Gaussian perturbation that makes FIVE a moving target for TextFooler. FIVE also delivers a solid performance against pre-made adversarial samples, which, together with its efficiently parallelizable approach, makes it attractive for further research.

\subsection{Conclusion}

We show that BERT-based classifiers can be hardened against both pre-made adversarial samples and active attack by a mechanism like TextFooler. The price for such improvements is some loss in accuracy on non-adversarial samples ranging from insignificant, as seen in the performance of our co-tuned model against pre-made samples, to substantial (10-20\%), as for the drops in regular benchmark metrics for our evaluation-time classifiers under attack by TextFooler. 

Future work should investigate the failure of FuSE to perform on the regular IMDB benchmark, combine some of our approaches into a single classifier, and tune the algorithm and implementation to allow for a longer hyperparameter search.

\pagebreak
\section*{Broader Impact}

BERT pre-trained classifiers opened the field of NLP classifiers to many practical applications, including sentiment and entailment classification. BERT-based classifiers score consistently high benchmark numbers \citep{devlin2018bert}. All that is left is the actual classification task, factoring out much of the daunting NLP aspect of such any text classification project. But can such classification be trusted, and to what degree? When text classification becomes usable, it also becomes tempting to use it as a replacement for human judgment. With popularity comes attack surface: Such classifiers were shown to be vulnerable to black-box trial-and-error antagonists like TextFooler \citep{jin2019bert}. These mechanisms use exhaustive search to produce adversarial examples that lead the classifier to the wrong result in over 90\% of samples examined, even when those samples were judged to be semantically equivalent by humans. Can classifiers be hardened against such attacks?

These questions need to be answered before we apply text classifiers to all kinds of applications as gatekeepers of civility and true representation of sentiments. Hardening classifiers against manipulation will protect meme-browsing youths and other vulnerable population segments from predatory individuals and trouble-seeking trolls, just as
it will protect intellectual property investments by thwarting manipulation of reviews and recommendations. 

\medskip

\bibliographystyle{apalike}

\bibliography{FireBERT}

\end{document}